\pdfoutput=1

\documentclass[11pt]{article}

\usepackage[final]{style/acl}

\usepackage{times}
\usepackage{latexsym}

\usepackage[T1]{fontenc}

\usepackage[utf8]{inputenc}

\usepackage{microtype}

\usepackage{inconsolata}

\usepackage{graphicx}       
\usepackage{multirow}       
\usepackage{subcaption}     
\usepackage{amssymb}        
\usepackage{bold-extra}     
\usepackage{bm}             
\usepackage[hang,flushmargin]{footmisc}  
\usepackage{booktabs}
\usepackage{tabularray}     
\NewColumnType{M}{Q[c,m]}
\NewColumnType{W}[1]{Q[l,m,wd=#1]}
\NewColumnType{L}{Q[l,m]}
\NewColumnType{N}[1]{Q[c,m,wd=#1]}
\usepackage{hyperref}
\newcommand{\LN}{\linebreak\noindent}    

\usepackage{pifont}

\usepackage[ruled, vlined, linesnumbered]{algorithm2e}
\usepackage{subfloat}
\usepackage{authblk}
\usepackage{supertabular}

\usepackage{enumitem}      
\setenumerate[1]{leftmargin=*}    
\setitemize[1]{leftmargin=*}      

\title{Automating PTSD Diagnostics in Clinical Interviews:\\ Leveraging Large Language Models for Trauma Assessments}

\author{
Sichang Tu,$^1$
Abigail Powers,$^1$
Natalie Merrill,$^1$
Negar Fani,$^1$\\
Sierra Carter,$^2$
Stephen Doogan,$^3$
Jinho D. Choi$^1$\\
$^1$Emory University, Atlanta, GA, USA\\
$^2$Georgia State University, Atlanta, GA, USA\\
$^3$Doogood Foundation, New York, NY, USA \\
{\small \texttt{\{sichang.tu, abigail.d.powers, natalie.merrill, nfani,jinho.choi\}@emory.edu}}\\
{\small \texttt{scarter66@gsu.edu}, \texttt{sdoogan@rlsciences.com}}
}

\begin{document}
\maketitle

\begin{abstract}

The shortage of clinical workforce presents significant challenges in mental healthcare, limiting access to formal diagnostics and services.
We aim to tackle this shortage by integrating a customized large language model (LLM) into the workflow, thus promoting equity in mental healthcare for the general population.
Although LLMs have showcased their capability in clinical decision-making, their adaptation to severe conditions like Post-traumatic Stress Disorder (PTSD) remains largely unexplored.
Therefore, we collect 411 clinician-administered diagnostic interviews and devise a novel approach to obtain high-quality data.
Moreover, we build a comprehensive framework to automate PTSD diagnostic assessments based on interview contents by leveraging two state-of-the-art LLMs, GPT-4 and Llama-2, with potential for broader clinical diagnoses. 
Our results illustrate strong promise for LLMs, tested on our dataset, to aid clinicians in diagnostic validation.
To the best of our knowledge, this is the first AI system that fully automates assessments for mental illness based on clinician-administered interviews.


\end{abstract}

\section{Introduction}
\label{sec:intro}

Mental health has become a vital element of overall well-being.
The prevalence of mental illness poses, however, a critical challenge to healthcare, underscoring the urgent need for an increased capacity of mental health services.
Only 29\% of people with psychosis receive formal care, leaving a significant portion completely untreated (WHO: \citet{WHO2020}).
Aside from obstacles such as high costs, limited awareness, and stigma surrounding mental health, the shortage of the mental health workforce has been a major factor exacerbating this gap.
According to WHO, the average ratio of mental health workers per 100,000 population was 13,\LN making it difficult for people to access reliable and readily administrated mental health diagnostics, as well as subsequent support and interventions.


\noindent The emergence of Large Language Models (LLMs) has suggested innovative solutions to this challenge.
Several studies have explored LLM applications in mental health for condition detection \cite{Zhang2022a}, support and counseling \cite{ma2023understanding} as well as clinical decision-making \cite{fu2023enhancing}, and shown the feasibility for LLMs to enhance the workforce of mental healthcare \cite{Hua2024a}.
By harnessing LLMs' ability to interpret languages that involve high expertise, it is possible to mitigate the service gap in the healthcare ecosystem through the automation of condition detection and diagnosis without the need of training so many professionals, which is both costly and time-consuming.


Despite these advancements, notable limitations persist in the current research on automatic diagnosis for mental health.
Most studies have focused on prevalent conditions like stress \cite{Lamichhane2023a} and depression \cite{qin2023read}, with scant attention to less common but more severe conditions like Post-traumatic Stress Disorder (PTSD).
Moreover, while prior studies have leveraged data from social media, clinical notes, and electronic health records, very few have utilized clinical interviews, and even in those cases, they rely on basic self-administered scales estimated in dialogues between computers and patients \cite{GalatzerLevy2023}.
No work has employed diagnostic interviews between real clinicians and patients that are systematically conducted, resulting in a dearth of practical research on\LN the automatic diagnosis of mental illness.


In this paper, we present an LLM-based system\LN that listens to long-hour conversations between clinicians and patients and performs diagnostic assessments for PTSD.
Our final model is evaluated by clinicians specialized in PTSD, suggesting a great potential for LLMs while highlighting certain limitations (Section~\ref{sec:Analysis}).
Our primary contributions are:\footnote{Our final model is publicly available through our open-source project at \url{https://github.com/emorynlp/TraumaNLP}.}


\begin{itemize}
    \item A new dataset comprising over 700 hours of interviews between clinicians and patients is created. Every interview consists of multiple diagnostic sections, featuring a series of questions and corresponding assessments from clinicians based on the interview contents (Section~\ref{sec:ptsd-data}).
    \item A novel and comprehensive pipeline is developed to process the interview dataset, so it can be used to build automatic assessment models on PTSD, which can be easily adapted to a broad range of diagnostic interviews (Section~\ref{sec:data-processing}).
    \item Assessment models achieving promising results are developed using two state-of-the-art LLMs, showcasing LLMs' ability to answer diagnostic questions through information extraction and text summarization on the interviews (Section~\ref{sec:experiments}).
\end{itemize}

\noindent To the best of our knowledge, this is the inaugural system designed to conduct diagnostic assessments\LN on mental health while interpreting real-world interviews administered by clinicians.
We believe that\LN this work will foster clinical collaboration between human experts and Artificial Intelligence, thus promoting equitable access to appropriate care for all populations affected by mental illness.

\section{Related Work}
\label{sec:related_work}

Pre-trained language models have been widely applied in many healthcare tasks \cite{Englhardt2023,Hu2023,Peng2023,Ma2023,Liu2023}.
The emergence of LLMs has introduced new capabilities and innovations in healthcare to this domain \citep{Nori2023,Cascella2023}.
This section introduces the related research of LLMs and their applications in healthcare, particularly in mental health.


\subsection{LLMs in Mental Health}

The advent of LLMs like GPT \citep{OpenAI2023}, Llama \citep{Touvron2023}, and PaLM \citep{Chowdhery2022a}
has sparked research into their applications in mental health \citep{Ji2023a}.
One key area is using conversational agents for mental health support and counseling,
where LLMs excel at generating empathetic responses \citep{Lai2023,ma2023understanding,loh2023harnessing},
highlighting their potential as digital companions or on-demand service providers.
Additionally, the research on decision-support systems for novice counselors underscores their potential to enhance mental healthcare provision \citep{fu2023enhancing}.


Research has also explored LLMs in disease detection and diagnosis \citep{Zhang2022a},
focusing on issues like depression \citep{qin2023read}, stress \citep{Lamichhane2023a},
and suicidality \citep{bhaumik2023mindwatch}.
Closer to our work, \citet{Bartal2023} use text-based narratives from new mothers to assess childbirth-related
PTSD with GPT and neural network models.
Although GPT showed moderate\LN performance, it holds promise for clinical diagnosis with further refinement.
These studies typically use zero/few-shot prompting for binary or multi-label classification,
demonstrating LLMs' capabilities in detecting mental health issues without fine-tuning,
despite challenges like unstable responses, potential bias, and interpretation inaccuracies.


Some research has pivoted towards fine-tuning LLMs for domain-specific performance enhancement.
\citet{Xu2023} present two fine-tuned models, Mental-Alpaca and Mental-FLAN-T5, outperforming GPT-3.5 and GPT-4 in multiple mental health prediction tasks.
Based on Llama-2, \citet{Yang2023c} train MentaLLaMA on 105K social media data enhanced by GPT.
The model performance is on par with other state-of-the-art methods, while providing interpretable analysis.

\subsection{LLMs in Clinical Interview and Diagnosis}

Research on using LLMs on clinical interview data and diagnosis is limited.
\citet{Wu2023} utilize GPT to augment the Extended Distress Analysis Interview Corpus by generating a new dataset from provided profile and rephrasing existing data.
The augmented data outperforms the original imbalanced data in PTSD diagnosis.
\citet{GalatzerLevy2023} adopt Med-PaLM-2 to predict Major Depression Disorder (MDD) and PTSD
on eight item Patient Health Questionnaire and PTSD Checklist-Civilian version ratings.

\begin{table*}[htbp!]
  \centering \small{ 
    \begin{tblr}{
      colspec = {MMMW{7.5cm}M},
      hline{1, Z} = {1pt, solid},
          hline{2} = {0.7pt, solid}
        }
      \bf Section & \bf Questions & \bf Variables & \SetCell{c}\bf Example Question & \bf Example Variable\\
      \tt LBI & 31  & 15 & What has been your primary source of income over the past month? & \texttt{lbi\_a1}\\
      \tt THH & 39 & 20 & In the past, have you been treated for any emotional or mental health problems with therapy or hospitalization?& \texttt{thh\_tx\_yesno}\\
      \tt CRA & 17 & 20 & What would you say is the one that has been most impactful where you are still noticing it affecting you?& \texttt{critaprobenotes}\\
      \tt CAP & 241 & 92 & In the past month, have you had any unwanted memories of the [Event] while you were awake, so not counting dreams?&{\texttt{dsm5capscritb01}\\\texttt{trauma1\_distress}}\\
    \end{tblr}}
  \caption{Statistics and examples for each of the four sections employed in this study.}
  \label{tab:qsets}
  \vspace{-1em}
\end{table*}

\section{PTSD Interview Data}
\label{sec:ptsd-data}

This study utilizes data from diagnostic interviews administered as part of a larger study on risk and resiliency to the PTSD development in a population seeking medical care \cite{gluck-2021}.
Participants were recruited from waiting rooms in primary care, gynecology and obstetrics, and diabetes medical clinics at a publicly funded, safety-net hospital.
Data were collected from 2012 to 2023, and inclusion criteria were ages between 18 and 65 with the capacity to provide informed consent.
The parent study was conducted according to the latest version of the Declaration of Helsinki \cite{helsinki-2013}, and consent from the participants was obtained after explaining the procedures.
The informed consent was approved by our Institutional Review Board and Research Oversight Committee.

\subsection{Participants}
\label{ssec:participants}

Participants were paid \$60.00 for this interview and underwent semi-structured diagnostic interviews conducted by doctoral-level clinicians or doctoral students supervised by a licensed clinical psychologist on staff.
A total of 411 interviews were conducted with 336 unique participants, some of whom had follow-up interviews after >1 month.
93.4\% of\LN the participants were women and 79.5\% were Black or African American ($M_{age}$ = 31.4), where 38.7\% had a high school education or less and 57.9\% reported a monthly household income of < \$1,000.

\subsection{Interview Procedures}
\label{ssec:interview-procedures}

The diagnostic interview begins with a section of the Longitudinal Interval Follow-Up Evaluation to assess global adaptive functioning across various psychosocial domains, including work, household, relationship as well as general functioning, and life satisfaction in the past month \cite{keller-1987}.
Videos of the interviews are recorded using online conferencing software such as Zoom and Microsoft Teams.
Each interview lasts 1.5 hours on average, involving the participant and 1-2 interviewers.

\subsection{Psychiatric Diagnoses and Treatment}
\label{ssec:sections}

A total of 10 sections are applied during the interview.
Among them, 4 sections are administered to the majority of participants; thus, this study focuses on those 4 sections.
The first two sections, the {Life Base Interview} (\texttt{LBI}) and the {Treatment History \& Health} (\texttt{THH}), are internally designed to assess the history of psychiatric diagnoses and treatment, as well as the presence of suicidality.
The other two sections, the {Criterion A} (\texttt{CRA}) and the {Clinician-Administered PTSD Scale for DSM-5} (\texttt{CAP}), follow the standard diagnostic criteria for PTSD outlined in the Diagnostic and Statistical Manual of Mental Disorders (DSM-5; \citet{weathers-2018}).
Every\LN section is accompanied by a set of questions, linked to variables that store pertinent values derived from the corresponding answers.
Table~\ref{tab:qsets} shows statistics and examples for each of the 4 sections.\footnote{Descriptions of all 10 sections are provided in Appendix~\ref{app:all-section-types}.}


\paragraph{\texttt{LBI}} It assesses the participant's functioning over the past month, addressing topics such as daily life, work, relationships with friends and family, and overall life satisfaction.

\paragraph{\texttt{THH}} It covers the participant's treatment/health history, including past physical and mental conditions as well as treatments received, such as medication and therapeutic services.

\paragraph{\texttt{CRA}} It assesses whether the participant has been exposed to (threatened) death, serious injury, or sexual violence, with a focus on potential traumatic experiences the participant might have endured.

\paragraph{\texttt{CAP}} It centers on issues the participant may have encountered due to traumatic events, including distress, avoidance of trauma-related stimuli, negative thoughts and feelings, and trauma-related arousal.

\section{Data Processing}
\label{sec:data-processing}

Every video is converted into an MP3 audio file and \textit{transcribed} by two automatic speech recognizers, whose results are \textit{aligned} to produce a high-quality transcript.
The transcript is \textit{segmented} into multiple sections based on the relevant questions, and each question is \textit{paired} with its assessment result.

\subsection{Transcription}
\label{ssec:transcription}

Two commercial tools, Rev AI\footnote{Rev AI: \url{https://www.rev.ai}} and Azure Speech-to-Text\footnote{Azure Speech-to-Text: \url{https://bit.ly/42r24pA}}, and an open-source tool, OpenAI Whisper \cite{radford-whisper-2023}, are tested for automatic speech recognition (ASR) on our dataset.
Whisper gives the lowest Word Error Rate (\texttt{WER}; \citet{klakow-wer-2002}) of 0.13, compared to 0.21 and 0.16 from Rev AI and Azure, respectively.
Whisper also exhibits better performance in handling noisy environments and numbers that Azure often misses or inaccurately transcribes (Table~\ref{tab:trans_number}).
Despite its superior ASR performance, Whisper does not identify speakers, a feature found in the others.
Thus, both Azure and Whisper are run on all audios and their results are combined to obtain the best outcomes.


\begin{table}[htbp!]
  \centering \small{ 
    \begin{tblr}{
      colspec = {MX},
      hline{1, Z} = {1pt, solid},
          hline{2} = {0.7pt, solid}}
      \bf Tool & \bf Examples \\
      \bf Azure & {(1) I got \textcolor{red}{2020} on the \textcolor{red}{24} with \textcolor{violet}{three}. \textcolor{violet}{Three} will be \textcolor{violet}{3} is turning \textcolor{olive}{2116}, \textcolor{olive}{one 15211}.\\(2) They happened in \textcolor{purple}{2017} and I'll be \textcolor{teal}{60} next month, so \textcolor{orange}{5556} something like that.}\\
      \bf Whisper & {(1) I got \textcolor{red}{two} to be \textcolor{red}{20} on the \textcolor{red}{24th}, well, \textcolor{violet}{three}, \textcolor{violet}{three} is turning \textcolor{olive}{20}, \textcolor{olive}{one 16}, \textcolor{olive}{one 15}, \textcolor{olive}{two 11}.\\(2) That happened in \textcolor{purple}{2017} and I'll be \textcolor{teal}{60} next month, so. \textcolor{orange}{55}, \textcolor{orange}{56}, something like that.}\\
    \end{tblr}}
  \caption{Comparisons between Azure and Whisper transcripts, with equivalent tokens coded in matching colors.}
  \label{tab:trans_number}
\vspace{-1.8em}
\end{table}

\subsection{Alignment}
\label{ssec:alignment}

To map the speaker diarization (SD) output from Azure to the Whisper output, {Align4D}\footnote{Align4D: \url{https://github.com/emorynlp/align4d}} is used such that
the first and last words of every utterance in the Azure output are aligned to their corresponding words in the Whisper transcript with speaker info, and form a speaker turn spanning all words between those words.
Some words in the Whisper transcript may get left out from this mapping, which are combined with either preceding or following adjacent utterances using heuristics.

Text-based Diarization Error Rate (\texttt{TDER}; \citet{gong-et-al-ictai-2023}) is used, more suitable than traditional metrics like \texttt{WER} or Diarization Error Rate (\texttt{DER}; \citet{fiscus-der-2006}), for evaluating text-based SD.
Transcripts from 29 audios produced by Microsoft Teams are used as the gold-standard, where Teams identifies speakers via different audio channels with near-perfect SD.
Our aligned method achieves a \texttt{TDER} of 0.56, a significant improvement over the \texttt{TDER} of 0.62 achieved by Azure alone.



\subsection{Segmentation}
\label{ssec:segmentation}

Each interview is conducted through multiple sections comprising a series of questions (Section~\ref{ssec:sections}), yet recorded as one continuous video.
It is crucial to segment the video into sections, each of which is split into sessions, where a session contains content relevant to a specific question.
Here, a session is defined as a list of utterances where the first utterance includes the corresponding question, and it is followed by another session whose first utterance includes the next question (if it exists).
Algorithm~\ref{alg:section-matching} describes how a section is matched in the transcript.

\vspace{-0.5em}
\renewcommand{\baselinestretch}{1.3}\selectfont
\begin{algorithm}
\small
  \caption{$\texttt{section\_match}(U, Q^c)$}
  \label{alg:section-matching}
  \KwIn{$U$: a list of utterances, $Q$: a list of questions.}
  \KwOut{An ordered list of tuples comprising utterance IDs and their matching scores.}

  $S \leftarrow \texttt{similarity\_matrix}(U, Q^c)$\;
  $T \leftarrow [\texttt{max}(S_{*,i}) : 1 \leq i \leq |Q^c|]$\;
  \lIf{
    $\texttt{average}(T) > 0.6\:\mathbf{and}$\\
    $\quad(|\texttt{select}(T, 0.8)| \geq 3 \:\mathbf{or}\: |\texttt{select}(T, 0.9)| \geq 2)$
   }{
    \KwRet{$\texttt{sequence\_alignment}(S)$}
   }
  \KwRet{$\varnothing$}
\end{algorithm}
\renewcommand{\baselinestretch}{1.0}\selectfont
\vspace{-0.5em}

\noindent Let $U$ be a list of utterances, and $Q^c$ a list of core questions for a specific section.\footnote{Core questions are required for retrieving essential information, while optional questions depend on the answers to the core questions, so are often skipped during the interview.}
$S \in \mathbb{R}^{|U| \times |Q^c|}$ is created, where $S_{i,j}$ is a similarity score between $u_i$ $\in U$ and $q_j \in Q^c$ (\texttt{L1}).
$T \in \mathbb{R}^{|Q^c|}$ is then created by selecting the maximum similarity score for every question (\texttt{L2}).
Given a function $\texttt{select}(T, s)$ that returns a list of scores in $T$ greater than $s$, the section is matched if $T$'s average score is > 0.6 (\texttt{L3}) and if there exist at least 3 or 2 questions whose matching scores are > 0.8 or 0.9, respectively (\texttt{L4}).
If the section is matched, \citet{gong-et-al-ictai-2023}'s sequence alignment algorithm is applied to $S$, which returns an ordered list of utterance IDs and their matching scores for questions in $Q^c$; otherwise, it returns an empty list (\texttt{L5}).
In our case, Sentence Transformer is used to create embeddings for utterances \& questions \cite{reimers2019sentencebert}, and cosine similarity is used to estimate the scores.

\noindent Overlap between spans of two sections may occur due to incorrect matching.
Algorithm~\ref{alg:remove-overlap} shows how to remove such overlaps.
Let $Q^c_i$ be a list of core questions for the $i$'th section, and $R_i = \texttt{sm}(U, Q^c_i)$ (\texttt{sm}: \texttt{section\_match}).
Given $(R_1, R_2)$, $R'_1$ is created by taking a subset of $R_1$ whose utterance IDs exist in $R_2$ (\texttt{L1}), and $R'_2$ is created similarity (\texttt{L2}).
If $R'_1$ contains more questions with scores > 0.6 than $R'_2$, implying $Q^c_1$ is more likely matched to the overlapped span than $Q^c_2$, $R'_2$ is removed from $R_2$ (\texttt{L4}); otherwise, $R'_1$ is removed from $R_1$ (\texttt{L5}).

\renewcommand{\baselinestretch}{1.3}\selectfont
\begin{algorithm}
\small
  \caption{$\texttt{remove\_overlap}(R_1, R_2)$}
  \label{alg:remove-overlap}
  \KwIn{$R_1$, $R_2$: ordered lists of tuples comprising utterance IDs and their matching scores for the first and second sections, respectively.}
  \KwOut{$(R_1, R_2)$: updated lists without overlaps.}

  $R'_1 \leftarrow [(i, s) : \forall {(i, s) \in R_1} \land (i, *) \in R_2]$\;
  $R'_2 \leftarrow [(i, s) : \forall {(i, s) \in R_2} \land (i, *) \in R_1]$\;
  \If{$|\texttt{select}(R'_1, 0.6)| > |\texttt{select}(R'_2, 0.6)|$}
    {\KwRet{$(R_1, R_2 \setminus R'_2)$}}
  {\KwRet{$(R_1 \setminus R'_1, R_2)$}}
\end{algorithm}
\renewcommand{\baselinestretch}{1.0}\selectfont

Finally, Algorithm~\ref{alg:map} shows how session spans are found for a specific section.
$C^e$ is a list of tuples comprising utterance IDs and their scores for the $k$'th section created by Algorithms \ref{alg:section-matching} and \ref{alg:remove-overlap} (\texttt{L1}) (\texttt{ro}: \texttt{remove\_overlap}).
$C^\ell$ is created in the same manner, except adapting the Levenshtein Distance (LD) as the similarity metric (\texttt{L2}) \cite{levenshtein1966binary}.
$\texttt{sel}(C, s)$ returns a list of tuples comprising utterance IDs and their matched question IDs, where the scores > $s$.
$\texttt{last}(U, Q^c_*)$ returns the first utterance ID of the $(k+1)$'th section if exists; otherwise, it returns the last utterance ID of $U$.
$C$ is created by taking the intersection of $C^e$ and $C^\ell$ whose scores > 0.8 and 0.7, and the last utterance ID (\texttt{L3}).\footnote{Any section not matched by Algorithm~\ref{alg:section-matching} is considered absent.}

For each span $U'$ of utterances between $C_i$ and $C_{i+1}$ (exclusive for both ends), a list $Q'$ of optional questions related to $C_i$ is created (\texttt{L5-7}).
$T^e$ is a list of tuples comprising utterance IDs in $U'$ and their matched question IDs in $Q'$ with scores > 0.8, and $T^\ell$ is created using LD (\texttt{L8-9}).
The intersection of $T^e$ and $T^\ell$ is appended to a list $O$ (\texttt{L10}), which is then merged with $C$ and sorted to produce $V$ (\texttt{L11}).

For each span $U''$ between $V_i$ and $V_{i+1}$, a list $Q''$ of any questions have not been matched in that span\LN is created (\texttt{L14}).
Bipartite matching bw. $U''$ and $Q''$\LN are performed to find matches optimizing several\LN criteria in Appendix~\ref{appendix:matching-criteria} (\texttt{L15}), accumulated, mer\-ged, and sorted to produce the final list (\texttt{L16-17}).

\renewcommand{\baselinestretch}{1.3}\selectfont
\begin{algorithm}
\small
  \caption{$\texttt{session\_match}(U, Q^c_{1..4}, Q^o_{1..4}, k)$}
  \label{alg:map}
  \KwIn{$U$: a list of utterances, $Q^{c|o}_{1..4}$: lists of core$|$optional questions for the 1..4'th sections, $k$: the section index to segment sessions in.}
  \KwOut{$(R_1, R_2)$: updated lists without overlaps.}

  $C^e \leftarrow \texttt{ro}(\texttt{sm}^e(U, Q^c_k), \texttt{sm}^e(U, Q^c_{\forall j \neq k}))$\;
  $C^\ell \leftarrow \texttt{ro}(\texttt{sm}^\ell(U, Q^c_k), \texttt{sm}^\ell(U, Q^c_{\forall j \neq k}))$\;
  $C \leftarrow (\texttt{sel}(C^e, 0.8) \cap \texttt{sel}(C^\ell, 0.7)) \cup \texttt{last}(U, Q^c_{*})$\;
  $O \leftarrow \varnothing$\;
  \For{$i \leftarrow 1$ \KwTo $(|C| - 1)$}{
      $U' \leftarrow$ a list of utterances between $C_{i}$ and $C_{i+1}$\;
      $Q' \leftarrow$ a list of questions in $Q^o_k$ related to $C_{i}$\;
      $T^e = \texttt{sel}(\texttt{sm}^e(U', Q'), 0.8)$\;
      $T^\ell = \texttt{sel}(\texttt{sm}^\ell(U', Q'), 0.7)$\;
      $O \leftarrow O \cup (T^e \cap T^\ell)$\;
  }
  $(V, W) = (\texttt{sorted}(C \cup O), \varnothing)$\;
  \For{$i \leftarrow 1$ \KwTo $(|V| - 1)$}{
      $U'' \leftarrow$ a list of utterances between $V_{i}$ and $V_{i+1}$\;
      $Q'' \leftarrow$ a list of questions in $Q^c_k \cup Q^o_k$ that are between $V_{i}$ and $V_{i+1}$\;
      $T \leftarrow$ the best bipartite matching results between $U''$ and $Q''$ optimizing several criteria in \ref{appendix:matching-criteria}\;
      $W \leftarrow W \cup T$
  }
  {\KwRet{$\texttt{sorted}(V \cup W)$}}
\end{algorithm}
\renewcommand{\baselinestretch}{1.0}\selectfont
\setlength{\textfloatsep}{5pt}

\subsection{Assessment Pairing}
\label{ssec:assessment-pairing}

Answers to the questions are used to determine the values of the variables (Table~\ref{tab:qsets}), resulting in many-to-many relations between questions and variables (many-questions to one-variable is the most common case).
Our data comprises five variable types.
(1) \emph{Scale} assesses on an ordinal scale with ratings for intensity, severity, or likeness.
(2) \emph{Category} selects among binary choices or distinct class labels.
(3) \emph{Measure} captures various units such as duration, frequencies, and ages.
(4) \emph{Notes} are summarized texts documented by the interviewers.
(5) \emph{Rule} is calculated based on predefined rules derived from the other variable types.
Table~\ref{tab:mapped_data} shows the statistics of all variables for each section in our dataset.

\begin{table}[htbp!]
  \centering \resizebox{\columnwidth}{!}{
    \begin{tblr}{
      colspec = {MMMMMMM},
      hline{1, Z} = {1pt, solid},
          hline{2} = {0.5pt, solid},
          hline{3} = {0.7pt, solid}
        }
      \SetCell[r=2]{c}\bf Type &\SetCell[c=5]{c}\bf Variables &&&&& \SetCell[r=2]{c}\bf Count\\
      & \textbf{\texttt{LBI}} & \textbf{\texttt{THH}} & \textbf{\texttt{CRA}} & \textbf{\texttt{CAP}} & \bf Total & \\
      Scale & 7 & 1& 0 & 40 &48 & 9,722 \\
      Category &4 & 9 & 15 & 3 & 31  & 4,258 \\
      Measure & 2& 0 & 1& 24 &27 & 3,482 \\
      Notes &1 & 10 & 3 & 0 &14 &  1,146 \\
      Rule & 1& 0 & 1 & 25 &27 & 6,326 \\
    \end{tblr}}
  \caption{Statistics of the five types of variables. Examples of these variables are provided in Appendix~\ref{appendix:variable-examples}.}
  \label{tab:mapped_data}
\end{table}

\vspace{-1em}
\begin{table*}[htbp!]
  \centering \small{ 
    \begin{tblr}{
      colspec = {MX[l,m]},
      hline{1, Z} = {1pt, solid},
          hline{2} = {0.7pt, solid},
          hline{3-Y} = {0.3pt, solid},
          width = \textwidth
        }
      \bf VT  & \SetCell{c}{\bf Template} \\
      \bf S\&C & [\texttt{INTRO}]. Based on the patient's interview history, please determine \{\textit{keywords}\} that the patient \{\textit{symptom}\}. [\texttt{RETURN}]. [\texttt{REASON}]. The "answer" should be in the range \{\textit{range}\}.\{\textit{attributes}\} \\
      \bf M & [\texttt{INTRO}]. Based on the patient's interview history, please calculate \{\textit{keywords}\} that the patient have \{\textit{symptom}\}. [\texttt{RETURN}]. [\texttt{REASON}]. The "answer" should be \{\textit{type}\}. \\
      \bf N & [\texttt{INTRO}]. Based on the formatted data from patient's interview, please determine whether or not the formatted data includes this specified information \{\textit{single\_slot}\}. [\texttt{RETURN}]. The "reason" gives a brief explanation on whether the formatted data includes or omits the information. The "answer" should be either "yes" or "no", indicating the presence or absence of the information in formatted data. \\

    \end{tblr}}
  \caption{Instruction templates for \textbf{S}cale, \textbf{C}ategory, \textbf{M}easure, and \textbf{N}otes variables. VT: Variable type, [\texttt{INTRO}]: Imagine you are a professional clinician, [\texttt{RETURN}]: Return the answer as a JSON object with "reason" and "answer" as the keys, [\texttt{REASON}]: The "reason" should provide a brief justification or explanation for the answer.}
  \label{tab:system_msg}
\vspace{-1.2em}
\end{table*}

\section{Experiments}
\label{sec:experiments}

\subsection{Dataset}

\noindent The original data contains 411 interviews (Sec.~\ref{sec:ptsd-data}).
Whisper tends to generate irrelevant or repetitive sequences when prolonged silences occur, rendering about $\approx$20\% of the resulting transcripts unusable.
To address this issue, silence removal and noise cancellation techniques are applied, recovering $\approx$80\% of them.
Among the 393 successful transcripts, 322 of them have human assessments (\textsection\ref{ssec:assessment-pairing}), which are used to evaluate our approach (Table~\ref{tab:video_data}).

\begin{table}[htbp!]
  \centering\small{  
    \begin{tblr}{
          colspec = {MMMMM},
          hline{1, Z} = {1pt, solid},
          hline{2} = {0.7pt, solid},
      }
                   & \textbf{Audios} & \textbf{Hours} & \textbf{Turns} & \textbf{Tokens} \\
        Original   & 411                 & 703             & 116,501        & 6,035,027          \\
        Transcribe & 393                 & 651             & 90,174        & 5,499,662          \\
        Evaluation & 322                 & 515             & 71,412        &4,335,977 \\
    \end{tblr}}
  \caption{Statistics of our PTSD interview dataset.}
  \label{tab:video_data}
\vspace{-0.7em}
\end{table}

\noindent Compared to other interview datasets\footnote{Statistics of the comparison is provided in Appendix~\ref{appendix:dataset_comparison}.},
our dataset is the largest in the mental health domain.
While existing datasets often involve human-machine dialogues or crowdworker simulations, ours consists of formal diagnostic interviews conducted entirely by clinicians,
making it the first clinician-administered interview dataset.
Additionally, our dataset aims to generate comprehensive diagnostic reports rather than just single scores, providing more detailed resource for clinical practice.

\subsection{Large Language Models (LLMs)}
\label{subsec:prompt}

The state-of-the-art commercial and open-source large language models, GPT-4 and Llama-2 \cite{Touvron2023}, are adapted for our experiments.\footnote{Specific versions, parameters, and costs for these large language models are provided in Appendix~\ref{appendix:experiment_cost} and \ref{appendix:llm-parameters}.}
For each question, a model takes all sessions related to the variable to which the question pertains (\textsection\ref{ssec:assessment-pairing}), and an instruction to provide the answer and explanation.
Table~\ref{tab:system_msg} shows our templates including replaceable patterns to generate the instruction for each variable type.
For \textbf{S}cale, \{\textit{keywords}\} can be replaced with "\textit{how severe}", and \{\textit{symptom}\} with "\textit{have unwanted dreams in the past month}".
For \textbf{C}ategory, \{\textit{keywords}\} can be replaced with "\textit{which of the following categories best describes}", and \{\textit{symptom}\} with "\textit{usual employment status}".
To constrain the answer generated by the model, details such as the answer \{\textit{range}\} for \textbf{S\&C}, and the value \{\textit{type}\} for \textbf{M}easure are incorporated.
\textbf{S} has a special pattern \{\textit{attributes}\}, directing the model to return a particular score under certain conditions.

Assessing model performance for \textbf{N}otes poses a challenge as they must be compared against text summarized by interviewers.
Given the complexity of this task, it is decomposed into multiple subtasks of binary classifications, information extraction, and categorization by adapting Chain-of-Thought \cite{wei2023chainofthought}.
First, GPT is asked to generate a list of slots for each \textbf{N} variable, based on a batch of summary notes from interviewers.
Because many of these slots have similar meanings, albeit varying in naming, GPT is again asked to cluster them.
The clusters generated by GPT are manually refined, resulting in final grouped slots that cover 95+\% of the initial generation.
For each of these slots, an LLM is tasked with determining if relevant content for the slot is present in the provided sessions.\footnote{Appendix~\ref{appendix:prompt-examples} gives slot examples for Notes variables.}


\begin{table*}[htbp!]
  \centering \small{ 
  \begin{tblr}{
    colspec = {MMMMMMMMMM},
    hline{1, Z} = {1pt, solid},
    hline{2} = {0.5pt, solid},
    hline{3} = {0.7pt, solid}}
    \SetCell[r=2]{c}\bf Type & \SetCell[r=2]{c}\bf Count & \SetCell[c=2]{c}\bf Accuracy & &\SetCell[c=2]{c}\bf RMSE && \SetCell[c=2]{c}\bf Bias && \SetCell[c=2]{c}\bf Recall &\\
    & & GPT-4 & Llama-2 & GPT-4 & Llama-2 & GPT-4 & Llama-2 & GPT-4 & Llama-2 \\
    Scale & 9,722 & 58.9 & 46.7 & 1.10 & 1.63 & -0.04 & 0.51 & - & -\\
    Scale$_g$ & 9,722 & 67.3 & 59.0 & 0.85 & 1.01 & -0.04 & 0.51 & -& -\\
    Category & 4,258 & 77.2 & 63.6 & - & - & -& - & - & -\\
    Measure & 3,482 & 64.4 & 56.5 & - & - & -0.34 & -0.004 & -& - \\
    Notes & 1,146 & -& - & -& - & -& - & 48.1& 52.7\\
    Rule & 6,326 & 68.4 & 59.8 & 0.80 & 0.92 & -0.15 & 0.44 & -& - \\
  \end{tblr}}
  \caption{Model performance on all variable types (\textsection\ref{ssec:assessment-pairing}) using four evaluation metrics (\textsection\ref{subsec:metrics}).}
  \label{tab:results}
\vspace{-1em}
\end{table*}


\subsection{Zero-shot V.S. Few-shot Settings}
\label{subsec:zero/few-shot}

Zero-shot and few-shot settings are tested across all variable types\footnote{Appendix~\ref{appendix:zero-few-shot} gives details on zero/few-shot settings.}.
For \textit{Scale}, two few-shot settings are explored: one including an example for a single scale point, and the other covering examples for all scale points.
For the GPT model, few-shot settings mostly outperform zero-shot settings in predicting \textit{Category}, \textit{Measure}, and \textit{Notes} variables.
For \textit{Scale}, the few-shot setting with a single example results in the lowest performance.
On the other hand, the few-shot setting including examples for all scale points shows a slight improvement in model performance.
Thus, few-shot settings are used for all experiments\LN with GPT.
In contrast, the Llama model consistently yields inferior outcomes with few-shot settings compared to zero-shot settings, leading us to adopt zero-shot settings for all Llama experiments.




\subsection{Evaluation Metrics}
\label{subsec:metrics}

Since each variable type is uniquely defined, different evaluation metrics are employed accordingly.
Accuracy is computed for all types except \emph{Notes}.
For \emph{Notes}, since the model identifies the presence of information in the provided sessions based on predefined slots, Recall is used as the primary metric to gauge the coverage of relevant information detected by the model.
For \emph{Scale}, the Root Mean Square Error (RMSE) and Bias evaluation are used.
RMSE quantifies the magnitude of errors, whereas Bias evaluation calculates the proportion of positive and negative residuals, thereby revealing any directional bias in the model predictions.


\subsection{Results}
\label{ssec:results}

Table~\ref{tab:results} gives the results for each variable type.
For \emph{Scale}, additional evaluation is conducted for \texttt{CAP} whose original scaling ranges from 0 to 4 where 0 indicates the absence of symptoms, 1 denotes minimal symptoms, and 2+ are considered symptoms that meet or exceed the threshold for clinical significance.
To reflect this clinical demarcation, scale points are categorized into three scale groups, 0, 1, and 2+, and evaluated as \textit{Scale}$_g$.\footnote{Appendix~\ref{appendix:model-performance-sections}/\ref{appendix:var_results} presents results for each section/variable.}

GPT consistently shows significantly higher accuracy, averaging 10.5\% more across all types than Llama, and reaches an accuracy of 68.4\% for \textit{Rule} accumulating outcomes of other types.
Regarding RMSE, GPT exhibits an error rate of 0.8 for \textit{Rule} using results of \textit{Scale}, implying that it is less than one scale off from human judgment on average.
In terms of Bias, ranging from -1: \textit{completely biased to negative} to 1: \textit{completely biased to positive}, GPT displays a marginal bias toward negative for \textit{Scale}, while Llama shows a strong positive bias, implying\LN that GPT is a bit conservative in predicting a higher scale, whereas Llama tends to overestimate.
GPT underestimates more than Llama for \textit{Measure}, however, showing a slight negative bias of 0.15 for \textit{Rule}.
For \textit{Notes}, Llama exhibits better performance with a recall of 52.7\% than GPT, suggesting that Llama is more effective in retrieving relevant information.
Considering that these models are not fine-tuned on\LN our data, this level of performance is very promising, as we can achieve a robust model for practical use with further training.

\section{Error Analysis}
\label{sec:Analysis}

\begin{table*}[htbp!]
    \centering \small{ 
        \begin{tblr}{
            colspec = {MX[l,m]MMMM},
            hline{1, Z} = {1pt, solid},
                    hline{2} = {0.5pt, solid},
                    hline{3} = {0.7pt, solid},
                    hline{4-8} = {0.3pt, solid},
                    width = \textwidth
                }
            \SetCell[r=2]{c}\bf Type & \SetCell[r=2]{c}\bf History                                                                                                                                                                                                                                                                                                                                                                                                                                                                                                                                                      & \SetCell[r=2]{c}\bf Gold & \SetCell[c=2]{c}\bf Auto  &   & \SetCell[r=2]{c}\bf Ext  \\
                                     &                                                                                                                                                                                                                                                                                                                                                                                                                                                                                                                                                                                  &                          & \bf GPT                            & \bf LM \\
            \bf MR                   & \textcolor{blue}{Have you had any physical reactions when something reminded you of what happened?} ... \textcolor{black}{I had a horrible headache.} ... \textcolor{blue}{How many times in the past month has that happened?} ... \textcolor{black}{Those two times.} ... \textcolor{blue}{How long did it take you to sort of feel back to normal?} \textcolor{black}{I swear. It took me a minute. I got up. I got a glass of water. It took me about. I say two to three hours.} ... \textcolor{blue}{So how bad was that Headache? Do you think there are any other symptoms?} \textcolor{black}{\textcolor{red}{It was extremely}. I never had. I had it like that.} & 4                        & 3                              & 2   \\
            \bf FN                   & ... \textcolor{blue}{can you think about like how often that might happen in the last month about?} \textcolor{black}{I feel like about like \textcolor{red}{five times a week}.}                                                                                                                                                                                                                                                                                                                                                                                                                                    & 5                        & 20                             & 20  & \checkmark \\
            \bf EI                   & ... \textcolor{blue}{when did those start for you?} ... \textcolor{black}{So, \textcolor{red}{since around age 12}, at least yeah yeah because it took me a long time to really trust my stepfather.}                                                                                                                                                                                                                                                                                                                                                                                                                  & 480                      & NA                              & 108 & \checkmark \\
            \bf TE                   & ... \textcolor{blue}{how satisfied and fulfilled have you felt about your life, with zero being like not at all, couldn't have a worse life, and 10 being perfect, couldn't have a better life?} \textcolor{black}{I would say a \textcolor{red}{C}, because it's a lot more things that I want to do to be at a 10.}                                                                                                                                                                                                                                                                                                & 2                        & 3                              & 3 &  \checkmark\\
            \bf SM                   & \textcolor{blue}{So how many times in the past month would you say some things made you upset that reminded you of it?} \textcolor{red}{Rarely, maybe like two, three times? Very rarely.}                                                                                                                                                                                                                                                                                                                                                                                                                          & 2                        & 1                              & 1 &  \checkmark\\
            \bf CR                   & ... \textcolor{blue}{thinking about your work in the past month, how have you been doing?} ... \textcolor{black}{It's a normal, consistent, um, \textcolor{red}{it's a normal, consistent routine} where I do the same thing, do the same thing every day.}                                                                                                                                                                                                                                                                                                                                                           & 40                       & NA                              & 40 &  \\
        \end{tblr}}
    \caption{Examples of the six error types. MR: Misaligned Reasoning, FN: False Negative, EI: External Information, TE: Transcription Error, SM: Session Mismatching, CR: Commonsense Reasoning. Gold: clinician's answers, Auto: model-predicted answers.  Ext: errors caused by external factors, not LLMs. NA: the model predicts \texttt{None}. Clinician's questions are highlighted in blue. Patient's key information to the questions are highlighted in red.}
    \label{tab:errors}
\vspace{-1em}
\end{table*}

A thorough error analysis is conducted by proportionally sampling 100+ examples per variable type.
Six types of major errors are identified (Table~\ref{tab:errors}), with only two attributed to LLMs and the remainder caused by external factors, implying that the true LLM performance may be even higher.


\paragraph{Misaligned Reasoning}

One predominant error type occurs when models deviate from instructions of the rating scheme, presenting seemingly logical reasoning, although it ultimately leads to incorrect conclusions.
In Table~\ref{tab:errors}, both models fail to align the key term provided by the participant, \textit{extremely}, with the definition of score 4 - ``Extreme, dramatic physical reactivity''. 
Llama tends to deviate further than GPT, resulting in a higher RMSE.


\paragraph{False Negatives} is a major error type caused by:
\begin{enumerate}
\setlength\itemsep{-0.1em}
\item Inaccurate assessments by clinicians. In Table~\ref{tab:errors}, the participant reports \textit{five times a week}, yet the clinician incorrectly records the frequency of monthly basis as \textit{5}, which should have been \textit{20} times a month.
\item Ambiguity in \textit{Scale} where answers may fall between two scales, resulting in potentially valid model predictions being marked incorrect.
\item The model's inability to recognize paraphrased information in \textit{Notes}, mistakenly indicating the absence of slot information. This issue particularly affects GPT's performance due to its strict interpretation of wording variations.
\end{enumerate}


\paragraph{External Information}

One common issue is the absence of external information, such as the prior knowledge about the patient (e.g., medical histories, demographics) or the content of previous interview questions.
In Table~\ref{tab:errors}, although both models see the onset of symptoms at age \textit{12}, they fail to provide an accurate response of the total symptom duration in months because the patient's current age (that is 52) is not provided in the transcript.
In this case, GPT tends to generate a \texttt{None} answer, while Llama tends to hallucinate the patient's age, and thus produces an answer based on an arbitrary assumption.
\vspace{-0.3em}


\paragraph{Transcription Error}

Transcription errors from automatic speech recognizers often cause LLMs to incorrectly interpret the answers, especially with short responses (e.g., \textit{yes}, \textit{no}, single digits like \textit{6}), medical terminologies, or non-verbal cues such as nodding. In Table~\ref{tab:errors}, the number `6' is incorrectly transcribed as `\textit{C}' in the participant's response.
\vspace{-0.3em}


\paragraph{Session Mismatching}

A question can be mismatched with the transcript, especially when the clinician extensively paraphrases it.
In such cases, the segmented session may or may not contain all the necessary information to answer the question.
In Table~\ref{tab:errors}, both models correctly answer based on the patient's response (1: Minimal).
However, due to the mismatch, the session is missing a part where the patient also indicates 2 (clearly present but still manageable), which is recorded as gold.


\paragraph{Commonsense Reasoning}

The models' limitations extend to inferring basic human experiences. Unable to deduce standard working hours from \textit{a normal, consistent routine} in Table~\ref{tab:errors}, the models fall short of clinician-like assumptions of a typical 40-hour workweek, showcasing a gap in applying commonsense logic to the assessment.


\section{Conclusion}
\label{sec:conclusion}

In this study, we undertook the task of automating PTSD diagnostics using 411 clinician-administered interviews.
To ensure the data quality, we develop an end-to-end pipeline streamlining transcription, alignment, segmentation, and assessment pairing.
We also construct a pioneering framework for this\LN task by leveraging two state-of-the-art LLMs.
Our findings reveal the substantial potential of LLMs in assisting clinicians with diagnostic validation and decision-making processes.
Our error analysis suggests future directions for improvement, such as incorporating external information or common-sense knowledge to engineer more comprehensive instructions. 
We envision that this framework holds promise for addressing a broader spectrum of mental health conditions and offers novel insights into LLM applications within the mental health domain.\LN
We plan to collect more data and train a custom LLM to better preserve patients' privacy, and develop a dialogue system to conduct the interviews.

\section*{Limitations}

Although the experiment results prove the capability of LLMs to automate PTSD diagnosis,
their applications in real-world unsupervised clinical settings are premature.
To avoid the possible negative influence of model errors on the patients,
we recommend using this framework as a supportive tool for clinicians in diagnostics and decision-making.

It should be noted that the clinician annotated gold assessment data is not perfect,
which may affect evaluation accuracy.
However, this framework makes it easier to identify and refine the inaccuracies in the gold assessment data and thus improve its overall validity.
We leave the data augmentation as the next step of our future work.

In addition, the experiments in this paper utilize LLMs without fine-tuning.
One limitation is that we have little control over the model predictions.
The models, especially Llama-2, generate unexpected outputs that violate the instructions.
Furthermore, data privacy concerns restrict the use of models like GPT for clinical data.
To address these issues and enhance framework adaptability, future work will focus on developing more controllable, open-sourced models that guarantee data protection in line with clinical domain restrictions.

Due to strict Institutional Review Board (IRB) regulations concerning the confidentiality of real patient information, we are unable to release the dataset, even in an anonymized format. However, recognizing the importance of contributing to the research community, we are pleased to announce that we will release the framework utilized in our study. This, we believe, will facilitate further research and innovation, as our methodology is versatile and can be adapted to a wide array of mental health conditions, provided the requisite interview question sets and video/transcripts are available.

\section*{Ethical Considerations}

The diagnostic interview data used in this paper was collected with informed consent approved by the Institutional Review Board (IRB) and Research Oversight Committee.
The authors and clinicians involved in the research have passed Research, Ethics, Compliance, and Safety Training through Collaborative Institutional Training Initiative\footnote{\url{https://about.citiprogram.org}} (CITI Program).
For the use of LLMs, this study exclusively employs anonymized interviews,
ensuring the confidentiality and privacy of all participants.
All practices in this research adhere to the ACL Code of Ethics.


\bibliography{custom}

\cleardoublepage
\appendix
\section{Section Details}
\label{app:all-section-types}

Table~\ref{tab:lbi_ex} - \ref{tab:cap_ex} give examples for 4 core sections.
Each example includes the standard interview \textbf{Q}uestion,
the \textbf{V}ariable that the question belongs to,
and the example \textbf{S}essions between the \textbf{C}linician and the \textbf{P}articipant.

The Mini International Neuropsychiatric Interview (\texttt{MINI}) is a brief, structured diagnostic interview for diagnosing 17 major psychiatric disorders \citep{sheehan1998mini}.
We adopt 6 modules from MINI to assess conditions such as Major Depressive Episode (\texttt{MDE}), Mania \& Hypomania (\texttt{MH}), PTSD (past incidents), Psychosis Symptoms (\texttt{PS}), Substance Use Disorder (\texttt{SUD}), and Alcohol Use Disorder (\texttt{AUD}).
Table~\ref{tab:mini_ex} provides an example from the \texttt{MDE} module.

\begin{table}[htbp!]
  \centering \small{ 
    \begin{tblr}{
      colspec = {Q[l,m]X},
      hline{1, Z} = {1pt, solid},
          hline{4} = {1pt, solid},
          width = \linewidth
        }
      \SetColumn{gray9}  \bf Q & What has been your primary source of income over the past month?                                                                                                                                                                                                                        \\
      \SetHline{2}{white, 0.5pt}
      \bf V                    & \texttt{lbi\_a1}                                                                                                                                                                                                                                                                        \\
      \SetHline{2}{white, 0.5pt}
      \bf S                    & {\textbf{C}: You got to do it all over again. Are you working full time?                                                                                                                                                                                                                \\
      \textbf{P}: Yes.}                                                                                                                                                                                                                                                                                                  \\
      \bf Q                    & How would you rate your overall satisfaction on a scale of 1 to 10, with 1 being the best and 10 being the worst?                                                                                                                                                                       \\
      \SetHline{2}{white, 0.5pt}
      \bf V                    & \texttt{lbi\_e1}                                                                                                                                                                                                                                                                        \\
      \SetHline{2}{white, 0.5pt}
      \bf S                    & {\textbf{C}: In the past month, like how satisfied have you felt with your life? If we were doing like a scale of one to 10, one is like, it's the worst. This is the worst I've ever had in my life. 10 being like, this is, I'm living my best life. Living my life like it's golden. \\
          \textbf{P}: I actually feel like that now. I actually do. Cause until January 1st of this year, I had been unemployed the last two years.}
    \end{tblr}}
  \caption{Two examples of the \texttt{LBI} section.}
  \label{tab:lbi_ex}
\end{table}

\begin{table}[htbp!]
  \centering \small{ 
    \begin{tblr}{
      colspec = {Q[l,m]X},
      hline{1, Z} = {1pt, solid},
          hline{4} = {1pt, solid},
          width = \linewidth
        }
      \SetColumn{gray9}  \bf Q & Do you have any current physical health conditions?                                                                                                                                                                                               \\
      \SetHline{2}{white, 0.5pt}
      \bf V                    & \texttt{thh\_medicalcond}                                                                                                                                                                                                                         \\
      \SetHline{2}{white, 0.5pt}
      \bf S                    & {\textbf{C}: OK, so now we're going to move on to talking about your health and treatment history. Do you currently have, do you have any current physical health conditions? Did you say no? OK, I couldn't hear what you were saying. Go ahead. \\
      \textbf{P}: I have a skin condition called eczema.}                                                                                                                                                                                                                          \\
      \bf Q                    & In the past, have you been treated for any emotional/mental health problems with therapy or hospitalization?                                                                                                                                      \\
      \SetHline{2}{white, 0.5pt}
      \bf V                    & \texttt{thh\_tx\_yesno}                                                                                                                                                                                                                           \\
      \SetHline{2}{white, 0.5pt}
      \bf S                    & {\textbf{C}: In the past, have you been treated for any emotional or mental health problem with therapy or hospitalization?                                                                                                                       \\
          \textbf{P}: No. Yes.}
    \end{tblr}}
  \caption{Two examples of the \texttt{THH} section.}
  \label{tab:thh_ex}
\end{table}

\begin{table}[htbp!]
  \centering \small{ 
    \begin{tblr}{
      colspec = {Q[l,m]X},
      hline{1, Z} = {1pt, solid},
          hline{4} = {1pt, solid},
          width = \linewidth
        }
      \SetColumn{gray9}  \bf Q & Tell me a little bit more about what happend. \\
      \SetHline{2}{white, 0.5pt}
      \bf V                    & \texttt{trauma1whathappened}                  \\
      \SetHline{2}{white, 0.5pt}
      \bf S                    & {\textbf{C}:  OK, and what would that be?     \\
          \textbf{P}: My mom worked at the airport here in xxx. It was the food catering place. They put the food, made the food for the planes. When I was a child, every year, they would sponsor a day at xxx. They would go out there and barbecue. We took over the whole picnic area. You had free entrance to the park, plus tickets to do all the little fair games and all that good stuff. Having a good time. My mom asked my stepfather to go with us because he had a car. He said he didn't wanna go and he wasn't going nowhere. So my mom put us all on the bus. We drove the bus out there. When we came home, it was like 11 o'clock. Of course, we living in xxx. You know that bus ride was long. It was dark, dark when we got home and she had all three of her children with her. My mom unlocked the door, closed that door, the house was pitch black. That man shot down them steps at my mama and all three of her children five times.}
    \end{tblr}}
  \caption{An example of the \texttt{CRA} section.}
  \label{tab:cra_ex}

\end{table}

\begin{table}[htbp!]
  \centering \small{ 
    \begin{tblr}{
      colspec = {Q[l,m]X},
      hline{1, Z} = {1pt, solid},
          hline{4} = {1pt, solid},
          width = \linewidth
        }
      \SetColumn{gray9}  \bf Q & Tell me a little bit more about what happend.                                                                                                                                                                                    \\
      \SetHline{2}{white, 0.5pt}
      \bf V                    & \texttt{dsm5capscritb
      01trauma1\_distress}                                                                                                                                                                                                                                        \\
      \SetHline{2}{white, 0.5pt}
      \bf S                    & {\textbf{C}:  To this day, let's say over the past month. So since like the beginning of April, end of March, have you had unwanted memories of this event? Does it randomly pop into your mind at all? Like while you're awake? \\
      \textbf{P}: Well, actually my daughter's in an abusive relationship. So yes, I do think about it a lot. Every time I see her, all I think about is my mom. How she endured it.}                                                                             \\
      \bf Q                    & How often in the past month?                                                                                                                                                                                                     \\
      \SetHline{2}{white, 0.5pt}
      \bf V                    & \texttt{dsm5capscritc02trauma1\_num}                                                                                                                                                                                             \\
      \SetHline{2}{white, 0.5pt}
      \bf S                    & {\textbf{C}:  So in the last month, thinking about the things that you have tried to avoid, how often would you say you've done that?                                                                                            \\
          \textbf{P}: I guess every day. I don't know. I just, the most I've done is just, and me avoiding stuff is me just sitting here smoking and playing my video game. That avoids me from thinking about anything negative in my life. And I just try to avoid that.}
    \end{tblr}}
  \caption{Two examples of the \texttt{CAP} section.}
  \label{tab:cap_ex}
\end{table}

\begin{table}[htbp!]
  \centering \small{ 
    \begin{tblr}{
      colspec = {Q[l,m]X},
      hline{1, Z} = {1pt, solid},
          hline{4} = {1pt, solid},
          width = \linewidth
        }
      \SetColumn{gray9}  \bf Q & For the past two weeks, were you depressed or down, or felt sad, empty or hopeless most of the day, nearly every day?                                                                                                                                        \\
      \SetHline{2}{white, 0.5pt}
      \bf V                    & \texttt{miniv7\_mde\_c\_a1}                                                                                                                                                                                                                                  \\
      \SetHline{2}{white, 0.5pt}
      \bf S                    & {\textbf{C}:  I'm going to ask you some different questions. We're going to focus on the past two weeks right now. So for the past two weeks, did you feel depressed, down, sad, empty or hopeless for most of the day, almost every day the past two weeks? \\
      \textbf{P}: Um, no.}                                                                                                                                                                                                                                                                    \\
    \end{tblr}}
  \caption{An example of the \texttt{MINI} section.}
  \label{tab:mini_ex}
\end{table}

\cleardoublepage

\section{Data Preprocessing Details}

\subsection{Final Matching Criteria}
\label{appendix:matching-criteria}

The best bipartite result should follow the criteria:
\begin{itemize}
\setlength\itemsep{0em}
  \item All matching IDs need to be ascending.
  \item Only edges whose embedding cosine similarity > 0.4 are kept.
  \item Maximize: $y = \sum_{i=1}^{n} a_i \cdot x_i$, subject to $x_i \geq 0$, for $i = 1, \ldots, n$.
\end{itemize}

\noindent In our case, let $n=9$, with the following variables:

\begin{itemize}
\setlength\itemsep{0em}
  \item $x_1$: the sum of Sentence Transformer (ST) cosine similarity scores of all edges
  \item $x_2$: the sum of Levenshtein Distance (LD) similarity scores of all edges
  \item $x_3$: the average ST cosine similarity scores of all matched questions
  \item $x_4$: the average LD similarity scores of all matched questions
  \item $x_5$: the total number of matched core questions
  \item $x_6$: the total number of matched questions that take the maximum ST cosine similarity result
  \item $x_7$: the total number of matched questions that take the maximum LD similarity result
  \item $x_8$: the total number of matched core questions that take the maximum ST cosine similarity result
  \item $x_9$: the total number of matched core questions that take the maximum LD similarity result
\end{itemize}

And the coefficients are set as:
\begin{itemize}
\setlength\itemsep{0em}
  \item $a_1 = a_2 = 1$
  \item $a_3 = a_4 = 1$
  \item $a_5 = a_6 = a_7 = 0.1$
  \item $a_8 = a_9 = 0.2$
\end{itemize}



\subsection{Variable Examples}
\label{appendix:variable-examples}

Table~\ref{tab:scale_var} - \ref{tab:rule_var} show examples for each variable type.
Every example includes the \textbf{V}ariable name,
replaceable \textbf{P}atterns for prompt generation (Section~\ref{sec:experiments}),
answer \textbf{R}ange, and covered \textbf{Q}uestions.
Note that \emph{Measure}, \emph{Notes}, and \emph{Rule} variables do not have a predefined range.
And \emph{Rule} variables are calculated from the results of their \textbf{R}elated \textbf{V}ariables.

\begin{table}[htbp!]
  \centering \small{ 
    \begin{tblr}{
      colspec = {Q[l,m]X},
      hline{1, Z} = {1pt, solid},
          hline{2} = {0.7pt, solid},
          hline{3-Y} = {0.3pt, solid},
          width = \linewidth
        }
      \bf V & \texttt{dsm5capscritb01trauma1\_distress}                                                                          \\
      \bf P & {\{\textit{keywords}\}: how intense in the past month                                                              \\
      \{\textit{symptom}\}: unwanted memories of the traumatic event while awake                                                 \\
      \{\textit{attributes}\}:                                                                                                   \\
      - If the symptom only exists in dreams, the answer should be 0.                                                            \\
          - If the symptom is not perceived as involuntary and intrusive, the answer should be 0.

      }                                                                                                                          \\
      \bf R & { 0: None,                                                                                                         \\
      1: Minimal, minimal distress or disruption of activities                                                                   \\
      2: Clearly Present, distress clearly presented but still manageable, some disruption of activities                         \\
      3: Pronounced, considerable distress, difficulty dismissing memories, marked disruption of activities                      \\
      4: Extreme, incapacitating distress, cannot dismiss memories, unable to continue activities}                               \\
      \bf Q & {        In the past month, have you had any unwanted memories of it while you were awake, so not counting dreams? \\
      - How does it happen that you start remembering it?                                                                        \\
      --Are these unwanted memories, or are you thinking about it on purpose?                                                    \\
      - How much do these memories bother you?                                                                                   \\
      - Are you about to put them out of your mind and think about something else?                                               \\
      -- Overall, how much of a problem is this for you?                                                                         \\
      --- How so?}                                                                                                               \\
    \end{tblr}}
  \caption{An example of the \emph{Scale} variable. Questions start with - are optional questions that might be skipped based on the participant's response.}
  \label{tab:scale_var}
\end{table}

\begin{table}[htbp!]
  \centering \small{ 
    \begin{tblr}{
      colspec = {Q[l,m]X},
      hline{1, Z} = {1pt, solid},
          hline{2} = {0.7pt, solid},
          hline{3-Y} = {0.3pt, solid},
          width = \linewidth
        }
      \bf V & \texttt{lbi\_a1}                                                          \\
      \bf P & {\{\textit{keywords}\}: which of the following categories best describes, \\
      \{\textit{symptom}\}: usual employment status}                                    \\
      \bf R & {            1: Full-Time Gainful Employment                              \\
      2: Part-Time Gainful Employment (30 hours or less/week)                           \\
      3: Unemployed But Expected by Self or Others                                      \\
      4: Unemployed But Not Expected by Self or Others (e.g., physically disabled)      \\
      5: Retired                                                                        \\
      6: Homemaker                                                                      \\
      7: Student (Includes Part-Time)                                                   \\
      8: Leave of Absence Due to Medical Reasons (e.g., holding job; plans to return)   \\
      9: Volunteer Work - Full Time                                                     \\
      10: Volunteer Work - Part Time                                                    \\
      11: Other                                                                         \\
          888: N/A
      }                                                                                 \\
      \bf Q & {What has been your primary source of income over the past month?}        \\
    \end{tblr}}
  \caption{An example of the \emph{Category} variable.}
  \label{tab:category_var}
\end{table}

\begin{table}[htbp!]
  \centering \small{ 
    \begin{tblr}{
      colspec = {Q[l,m]X},
      hline{1, Z} = {1pt, solid},
          hline{2} = {0.7pt, solid},
          hline{3-Y} = {0.3pt, solid},
          width = \linewidth
        }
      \bf V & \texttt{dsm5capscritb01trauma1\_num}                                               \\
      \bf P & {\{\textit{keywords}\}: how intense in the past month                              \\
      \{\textit{symptom}\}: unwanted memories of the traumatic event while awake                 \\
      \{\textit{type}\}: an integer representing the frequency of the symptom in the past month} \\
      \bf Q & {- How often have you had these memories in the past month?}                       \\
    \end{tblr}}
  \caption{An example of the \emph{Measure} variable. The corresponding question for this question is optional, which might be skipped if the participant denies the presence of the symptom.}
  \label{tab:measure_var}
\end{table}

\begin{table}[htbp!]
  \centering \small{ 
    \begin{tblr}{
      colspec = {Q[l,m]X},
      hline{1, Z} = {1pt, solid},
          hline{2} = {0.7pt, solid},
          hline{3-Y} = {0.3pt, solid},
          width = \linewidth
        }
      \bf V & \texttt{critaprobenotes}                                                                                                                  \\
      \bf P & {\{\textit{slots}\}:                                                                                                                      \\
      - trauma\_reactions                                                                                                                               \\
      - trauma\_details                                                                                                                                 \\
      - life\_changes                                                                                                                                   \\
      - coping\_and\_changes                                                                                                                            \\
      - worldview\_changes                                                                                                                              \\
      - health\_concerns                                                                                                                                \\
      - family\_and\_social\_context                                                                                                                    \\
      - nightmare\_details                                                                                                                              \\
      - intrusive\_experiences                                                                                                                          \\
      - trauma\_cognition                                                                                                                               \\
      - trust\_and\_safety                                                                                                                              \\
      - impact\_assessment                                                                                                                              \\
      - age\_and\_time\_factors                                                                                                                         \\
      - substance\_use                                                                                                                                  \\
      - therapy\_and\_progress                                                                                                                          \\
          - eating\_disorders
      }                                                                                                                                                 \\
      \bf Q & {        You discussed a number of traumas in the last visit with our team members.                                                       \\
      What would you say is the one that has been most impactful where you are still noticing it affecting you?                                         \\
      -* How much do you think about what happened to this day?                                                                                         \\
      -* How often do you have nightmares about what happend?                                                                                           \\
      -* How much did it change the way you think about yourself and the world?                                                                         \\
      - In the past month, which of these have you thought about more often or had nightmares about or find yourself purposely avoiding thinking about? \\
      -- Are there any other stressors that you find yourself thinking about when you don't want to or avoiding?}                                       \\
    \end{tblr}}
  \caption{An example of the \emph{Notes} variable. Questions start with - are optional questions which might be skipped based on the participant's response. Questions start with * are recurrent questions which might be asked multiple times during the interview.}
  \label{tab:notes_var}
\end{table}

\begin{table}[htbp!]
  \centering \small{ 
    \begin{tblr}{
      colspec = {Q[l,m]X},
      hline{1, Z} = {1pt, solid},
          hline{2} = {0.7pt, solid},
          hline{3-Y} = {0.3pt, solid},
          width = \linewidth
        }
      \bf V  & \texttt{dsm5capscritb01trauma1}            \\
      \bf R  & { 0: Absent                                \\
      1: Mild/subthreshold                                \\
      2: Moderate/threshold                               \\
      3: Severe/markedly elevated                         \\
      4: Extreme/incapacitating}                          \\
      \bf RV & {\texttt{dsm5capscritb01trauma1\_distress} \\ \texttt{dsm5capscritb01trauma1\_num}}\\
    \end{tblr}}
  \caption{An example of the \emph{Rule} variable.}
  \label{tab:rule_var}
\end{table}





\newpage

\section{Experiments Details}

\subsection{Dataset Comparison}
\label{appendix:dataset_comparison}
Table~\ref{tab:dataset_compare} gives the comparison with related datasets in the metal health domain.

\begin{table}[htbp!]
  \centering \fontsize{8pt}{8pt}\selectfont 
  \begin{tblr}{
    colspec = {Q[c,m]Q[c,m]Q[c,m]Q[c,m]Q[c,m]},
    hline{1, Z} = {1pt, solid},
    hline{Y} = {0.5pt, solid},
    hline{2} = {0.5pt, solid},
    width = \linewidth
    }
    \bf Dataset & \bf A & \bf H & \bf Turns & \bf Utters \\
    {DAIC\\ \citep{DAIC}} & 189  & 51  & - & - \\
    {AViD\\ \citep{AViD}} & 300  & 240  & - & - \\
    {EATD\\ \citep{EATD}} & 162  & 2.26  & - & - \\
    {Psych8k\\ \citep{Liu2023a}} & 260  & 260  & - & - \\
    {D4\\ \citep{D4}} & -  & -  & 28,855 & 81,559 \\
    {ESConv\\ \citep{ESConv}} & -  & -  & - & 31,410 \\
    \bf Ours & \bf 322 & \bf 515 & \bf 71,412 & \bf 142,824 \\
  \end{tblr}
  \caption{Comparisons with existing mental health interview/dialogue datasets in terms of \textbf{A}udio counts, total \textbf{H}ours, total and utterances.}
  \label{tab:dataset_compare}
\end{table}

\subsection{Details on Zero-shot/Few-shot Settings}
\label{appendix:zero-few-shot}
We randomly sampled 30 instances for each variable type and asked both models to predict under zero-shot and few-shot settings.
For the GPT model, few-shot settings generally yield better performance.
However, the Llama model consistently fails to follow instructions as the context length grows, leading to significant degradation with few-shot prompting.
Additionally, we observed a 28\% increase in the likelihood of generating an unexpected response format, such as deviating from the requested JSON format, when using few-shot settings.

\begin{table}[htbp!]
  \centering \small{ 
  \begin{tblr}{
    colspec = {MMMMM},
    hline{1, Z} = {1pt, solid},
    hline{2} = {0.5pt, solid},
    hline{3} = {0.7pt, solid},
    }
    \SetCell[r=2]{c}\bf Type &  \SetCell[c=2]{c}\bf Zero-shot  && \SetCell[c=2]{c}\bf Few-shot &\\
    &  GPT-4 & Llama-2 & GPT-4 & Llama-2 \\
    Scale & 60.0 & 50.0 & 63.3 & 36.7\\
    Scale$_1$ & - & - & 56.7 & 40.0\\
    Category & 43.3  & 40.0 & 46.7 & 33.3\\
    Measure &  56.7 & 56.7 & 60.0& 50.0 \\
    Notes & 41.0 & 42.7 & 43.6 & 34.9\\
  \end{tblr}}
  \caption{Model performance on zero-shot and few-shot settings. Scale$_1$ refers to the few-shot setting that only include one example for a single scale point. Accuracy is the metric used for all variable types except Notes variables, which are evaluated using Recall.}
  \label{tab:zero-few}
\end{table}

\subsection{Experiment Costs}
\label{appendix:experiment_cost}
\paragraph{GPT-4}
The pricing of the GPT-4 Turbo model is \$0.01/1K tokens for input and \$0.03/1K tokens for output.
We spend approximately \$300 (upper bound) to complete GPT experiments in this paper.

\paragraph{Llama-2}
We use a single NVIDIA H100 GPU for Llama inferences with a batch size of 1,
taking roughly 10 seconds per request.
Completing a full set of experiments on all samples requires \textasciitilde 3 days.

\subsection{LLM Configurations}
\label{appendix:llm-parameters}

We utilize \texttt{gpt-4-1106-preview}, the latest GPT-4 Turbo model, and \texttt{llama-2-70b-chat-hf}, the largest Llama-2 model.
For GPT, to enhance the stability and consistency of the model output, we configure the \textit{temperature} parameter to 0.
This adjustment makes the model's response more deterministic.
Besides, we also employ parameters exclusive to GPT-4 Turbo and GPT-3.5 Turbo, namely \textit{response\_format} and \textit{seed}.
Setting \textit{response\_format} to "json\_object" constrains the model to generate parsable JSON strings, facilitating easier data handling and analysis.
Despite ChatGPT's non-deterministic nature, \textit{seed} parameter enables users to obtain consistent outputs across multiple requests, as long as there are no changes at the system level.

As for the Llama, we conduct experiments involving different \textit{temperature}, \textit{top\_p}, and \textit{repetition\_penalty} separately.
The results indicate that the model gives better performance with a \textit{temperature} setting of 0.3, a \textit{top\_p} of 0.9, and a \textit{repetition\_penalty} of 1.

\begin{table*}[htbp!]
  \centering \small{ 
    \begin{tblr}{
      colspec = {Q[l,m]X},
      hline{1, Z} = {1pt, solid},
          hline{2} = {0.7pt, solid},
          hline{3-Y} = {0.3pt, solid},
          width = \linewidth
        }
      \bf Step & \bf Template                                                                                                                                                                                                                                                                                                                                                                                                                                                          \\
      \bf NSG  & {As a clinician who has conducted interviews with multiple patients, you are tasked with structuring the interview data into a more organized format. To achieve this, identify general "slots" from the interview question and answers. These slots should represent key themes or types of information that can be adapted to various responses from different patients.                                                                                            \\For each identified slot, provide a brief explanation of why it has been chosen, focusing on its relevance and utility in categorizing interview data. \\Your findings should be presented in a JSON format as a list, for example: [\{"reason": "This slot captures the primary health concern of the patient, a common theme across all interviews", "slot": "primary\_health\_concern" \}, \{"reason": "This slot pertains to the patient's lifestyle habits, which is crucial for understanding health context", "slot": "lifestyle\_habits" \} ].\\Remember to ensure that the slots are broad enough to be applicable across different patient responses yet specific enough to offer meaningful categorization.} \\
      \bf NSM  & {Imagine you are a clinician who documents patient interviews in a structured, slot-filling manner. Sometimes, certain slots may have overlapping or similar content. Your task is to review a given list of slots and merge those that are similar. The merged results should be returned as a JSON object, where each key represents a merged slot, and the corresponding value is a list of the original slots that have been combined under this merged category. \\For instance, if the list of slots is: ["daily\_routine", "work\_events", "daily\_activity", "daytime\_activities", "work\_routine"], a possible merged result could be: \{"daily\_routine": ["daily\_routine", "daily\_activity", "daytime\_activities"], "work": ["work\_events", "work\_routine"]\}. \\When you receive a list of slots, analyze and merge them accordingly, ensuring that the merged slots are logically grouped and accurately represent the original information categories.} \\
      \bf NSF  & Imagine you are a professional clinician. Based on the patient's interview history, please extract specific information and fill in the following slots: \{slots\}. If the interview history does not provide information for any of these slots, please enter an empty string ('') for that slot. Return the answer as a JSON object.                                                                                                                                \\
    \end{tblr}}
  \caption{Prompts used for \textbf{N}otes Variable \textbf{S}lot structure \textbf{G}eneration, \textbf{M}erging, and \textbf{F}ormatting.}
  \label{tab:note_slot}
\end{table*}

\subsection{Slot Examples for Notes Variable}
\label{appendix:prompt-examples}

Table~\ref{tab:note_slot} outlines the process for generating, merging, and formatting the slots in \emph{Notes} variables (\textsection\ref{subsec:prompt}).
Initially, we compile all clinician-summarized notes for each \emph{Notes} variable and input them into the GPT model using the NSG prompt to produce a list of slots.
Due to potential overlaps, the NSM prompt directs the model to consolidate these slots into clusters,
ensuring both conciseness and comprehensiveness.
Subsequently, the NSF prompt is used to format both the gold-standard summaries and the corresponding interview sessions,
facilitating a straightforward comparison of the structured slot arrangements.

\subsection{Model Performance by Sections}
\label{appendix:model-performance-sections}
Table~\ref{tab:section_results} presents model performances by each section.
Note that \texttt{THH} section lacks \emph{Measure} and \emph{Rule} variables,
whereas \texttt{CRA} section does not contain \emph{Scale} variables.
The grouped scale$_g$ is exclusively applied within the \texttt{CAP} section.

\begin{table*}[htbp!]
  \centering \small{ 
    \begin{tblr}{
      colspec = {MMMMMMMMMMM},
      hline{1, Z} = {1pt, solid},
      hline{2} = {0.5pt, solid},
      hline{3, 8, 11, 15} = {0.7pt, solid}
        }
      \SetCell[r=2]{c}                                        & \SetCell[r=2]{c}\bf Type & \SetCell[r=2]{c}\bf Count & \SetCell[c=2]{c}\bf Accuracy &         & \SetCell[c=2]{c}\bf RMSE &         & \SetCell[c=2]{c}\bf Bias &         & \SetCell[c=2]{c}\bf Recall &         \\
                                                              &                          &                           & GPT-4                        & Llama-2 & GPT-4                    & Llama-2 & GPT-4                    & Llama-2 & GPT-4                      & Llama-2 \\
      \SetCell[r=5]{c} \rotatebox[origin=c]{90}{\texttt{LBI}} & Scale                    & 1,281                     & 54.6                         & 44.7    & 1.26                     & 1.42    & 0.46                     & 0.45    & -                          & -       \\
                                                              & Category                 & 594                       & 74.6                         & 67.3    & -                        & -       & -                        & -       & -                          & -       \\
                                                              & Measure                  & 99                        & 68.7                         & 66.7    & -                        & -       & -0.16                    & -0.09   & -                          & -       \\
                                                              & Notes                    & 203                       & -                            & -       & -                        & -       & -                        & -       & 42.0                       & 50.8    \\
                                                              & Rule                     & 215                       & 43.3                         & 37.7    & 0.94                     & 0.98    & 0.44                     & 0.43    & -                          & -       \\

      \SetCell[r=3]{c} \rotatebox[origin=c]{90}{\texttt{THH}} & Scale                    & 29                        & 55.2                         & 51.7    & 1.20                     & 1.25    & 0.23                     & 0.43    & -                          & -       \\
                                                              & Category                 & 1,527                     & 92.6                         & 85.9    & -                        & -       & -                        & -       & -                          & -       \\
                                                              & Notes                    & 633                       & -                            & -       & -                        & -       & -                        & -       & 52.5                       & 59.8    \\

      \SetCell[r=4]{c} \rotatebox[origin=c]{90}{\texttt{CRA}} & Category                 & 1,737                     & 63.7                         & 42.4    & -                        & -       & -                        & -       & -                          & -       \\
                                                              & Measure                  & 143                       & 63.6                         & 55.9    & -                        & -       & -0.58                    & -0.36   & -                          & -       \\
                                                              & Notes                    & 310                       & -                            & -       & -                        & -       & -                        & -       & 47.2                       & 43.5    \\
                                                              & Rule                     & 146                       & 91.8                         & 71.9    & 0.38                     & 0.97    & 0.83                     & -0.51   & -                          & -       \\

      \SetCell[r=5]{c} \rotatebox[origin=c]{90}{\texttt{CAP}} & Scale                    & 8,412                     & 59.6                         & 47.0    & 1.07                     & 1.66    & -0.14                    & 0.52    & -                          & -       \\
                                                              & Scale$_g$                & 8,412                     & 69.3                         & 61.2    & 0.77                     & 0.93    & -0.15                    & 0.53    & -                          & -       \\
                                                              & Category                 & 400                       & 81.0                         & 64.5    & -                        & -       & -                        & -       & -                          & -       \\
                                                              & Measure                  & 3,240                     & 64.2                         & 56.3    & -                        & -       & -0.33                    & 0.01    & -                          & -       \\
                                                              & Rule                     & 5,965                     & 68.8                         & 60.4    & 0.81                     & 0.92    & -0.19                    & 0.46    & -                          & -       \\
    \end{tblr}}
  \caption{Model performances on 4 sections (\textsection\ref{ssec:assessment-pairing}) using four evaluation metrics (\textsection\ref{subsec:metrics}).}
  \label{tab:section_results}
\end{table*}

\subsection{Model Performance by Variables}
\label{appendix:var_results}

Table~\ref{tab:var_results} lists results for each variable following the four evaluation metrics (Section~\ref{subsec:metrics}).

\onecolumn
{
  \renewcommand{\arraystretch}{1.5}
  \begin{center}
    \small
    \tablefirsthead{%
      \toprule
      \multicolumn{1}{c}{\multirow{2}{*}{\textbf{Variable}}} & \multicolumn{1}{c}{\multirow{2}{*}{\textbf{Count}}} & \multicolumn{2}{c}{\textbf{Acc}}                  & \multicolumn{2}{c}{\textbf{RMSE}}                 & \multicolumn{2}{c}{\textbf{Bias}}                 & \multicolumn{2}{c}{\textbf{Recall}}               \\ \cline{3-10}
      \multicolumn{1}{c}{}                                   & \multicolumn{1}{c}{}                                & \multicolumn{1}{c}{GPT} & \multicolumn{1}{c}{LM2} & \multicolumn{1}{c}{GPT} & \multicolumn{1}{c}{LM2} & \multicolumn{1}{c}{GPT} & \multicolumn{1}{c}{LM2} & \multicolumn{1}{c}{GPT} & \multicolumn{1}{c}{LM2} \\ \midrule
    }
    \tablehead{%
      \multicolumn{10}{l}{\small\sl continued from previous page}\\
      \toprule
      \multicolumn{1}{c}{\multirow{2}{*}{\textbf{Variable}}} & \multicolumn{1}{c}{\multirow{2}{*}{\textbf{Count}}} & \multicolumn{2}{c}{\textbf{Acc}}                  & \multicolumn{2}{c}{\textbf{RMSE}}                 & \multicolumn{2}{c}{\textbf{Bias}}                 & \multicolumn{2}{c}{\textbf{Recall}}               \\ \cline{3-10}
      \multicolumn{1}{c}{}                                   & \multicolumn{1}{c}{}                                & \multicolumn{1}{c}{GPT} & \multicolumn{1}{c}{LM2} & \multicolumn{1}{c}{GPT} & \multicolumn{1}{c}{LM2} & \multicolumn{1}{c}{GPT} & \multicolumn{1}{c}{LM2} & \multicolumn{1}{c}{GPT} & \multicolumn{1}{c}{LM2} \\ \midrule
    }
    \tabletail{%
      \bottomrule
      \multicolumn{10}{r}{\small\sl continued on next page}\\
    }
    \tablelasttail{\bottomrule}
    \bottomcaption{Model performances on all variable (\textsection\ref{ssec:assessment-pairing}) using four evaluation metrics (\textsection\ref{subsec:metrics}).}
    \begin{supertabular}{lccccccccc}
      \multicolumn{10}{c}{\bf \texttt{Scale} Variable}\\
      \hline
      lbi\_a2b                            & 199                        & 53.8                     & 37.7 & 1.31                       & 2.05 & 0.37                      & 0.68 &-&-\\
      lbi\_a3                             & 201                        & 56.7                     & 52.7 & 0.96                       & 0.99 & 0.31                      & 0.47 &-&-\\
      lbi\_a4                             & 63                         & 50.8                     & 55.6 & 0.90                       & 1.02 & 0.23                      & 0.29 &-&-\\
      lbi\_b1a\_family                    & 207                        & 46.9                     & 54.6 & 1.78                       & 1.36 & 0.49                      & 0.11 &-&-\\
      lbi\_b2                             & 212                        & 60.8                     & 44.8 & 1.19                       & 1.02 &   0.47                    & 0.49 &-&-\\
      lbi\_d                              & 194                        & 52.1                     & 38.7 & 1.29                       & 1.29 & 0.53                      & 0.53 &-&-\\
      lbi\_e1                             & 205                        & 58.5                     & 36.1 & 0.91                       & 1.66 &  0.65                     & 0.42 &-&-\\
      dx\_understanding                   & 29                         & 55.2                     & 51.7 & 1.20                       & 1.25 & 0.23                      & 0.43 &-&-\\
      {dsm5capscritb01trauma1\_distress}  & 257                        & 59.5                     & 53.3 & 0.89                       & 1.67 & -0.44                      & 0.48 &-&-\\
      {dsm5capscritb02trauma1\_distress}  & 254                        & 69.7                     & 53.9 & 0.72                       & 1.16 &   -0.56                    & 0.32 &-&-\\
      {dsm5capscritb03trauma1\_distress}  & 249                        & 67.9                     & 51.8 & 0.73                       & 1.35 &  -0.05                     & 0.68 &-&-\\
      {dsm5capscritb04trauma1\_distress}  & 259                        & 57.1                     & 40.5 & 0.96                       & 1.25 &   -0.35                    & 0.47 &-&-\\
      {dsm5capscritb05trauma1\_distress}  & 243                        & 63.8                     & 56.8 & 0.90                       & 1.04 &  -0.11                     & 0.37 &-&-\\
      {dsm5capscritc01trauma1\_distress}  & 253                        & 46.2                     & 39.9 & 1.77                       & 1.92 &   -0.34                    & 0.53 &-&-\\
      {dsm5capscritc02trauma1\_distress}  & 243                        & 58.0                     & 45.7 & 0.99                       & 1.20 &  -0.04                     & 0.64 &-&-\\
      {dsm5capscritd01trauma1\_distress}  & 242                        & 66.1                     & 53.7 & 0.92                       & 1.13 &  -0.10                     & 0.30 &-&-\\
      {dsm5capscritd02trauma1\_distress}  & 256                        & 56.6                     & 36.7 & 0.85                       & 1.31 & -0.06                      & 0.83 &-&-\\
      {caps5trauma1related\_d02}          & 164                        & 57.9                     & 55.5 & 0.97                       & 0.85 &  -0.71                     & 0.07 &-&-\\
      {dsm5capscritd03trauma1\_distress}  & 248                        & 61.7                     & 58.9 & 0.94                       & 0.92 &  -0.56                     & 0.24 &-&-\\
      {dsm5capscritd04trauma1\_distress}  & 252                        & 56.0                     & 49.2 & 0.93                       & 1.13 &    -0.03                   & 0.55 &-&-\\
      {caps5trauma1related\_d04}          & 160                        & 63.8                     & 54.4 & 0.89                       & 0.84 &   -0.28                    & 0.10 &-&-\\
      {dsm5capscritd05trauma1\_distress}  & 253                        & 57.7                     & 47.8 & 1.00                       & 1.18 & -0.08                      & 0.53 &-&-\\
      {caps5trauma1related\_d05}          & 138                        & 53.6                     & 44.9 & 1.06                       & 0.96 &  -0.56                     & 0.21 &-&-\\
      {dsm5capscritd06trauma1\_distress}  & 255                        & 53.5                     & 47.5 & 1.01                       & 1.23 &   0.09                    & 0.66 &-&-\\
      {caps5trauma1related\_d06}          & 156                        & 51.3                     & 41.0 & 0.98                       & 0.90 &  -0.47                     & 0.35 &-&-\\
      {dsm5capscritd07trauma1\_distress}  & 257                        & 59.5                     & 45.5 & 0.88                       & 1.22 &  0.04                     & 0.67 &-&-\\
      {caps5trauma1related\_d07}          & 128                        & 55.5                     & 44.5 & 0.96                       & 0.94 &  -0.16                     & 0.35 &-&-\\
      {dsm5capscrite01trauma1\_distress}  & 257                        & 60.3                     & 46.7 & 0.79                       & 1.13 &   0.33                    & 0.78 &-&-\\
      {caps5trauma1related\_e01}          & 148                        & 52.7                     & 33.8 & 3.54                       & 3.44 &   -0.74                    & 0.06 &-&-\\
      {dsm5capscrite02trauma1\_distress}  & 251                        & 67.3                     & 61.0 & 0.71                       & 1.11 &  0.02                     & 0.31 &-&-\\
      {caps5trauma1related\_e02}          & 50                         & 74.0                     & 58.0 & 1.09                       & 1.26 &  -0.38                     & 0.43 &-&-\\
      {dsm5capscrite03trauma1\_distress}  & 255                        & 51.4                     & 47.1 & 1.09                       & 1.20 & 0.32                      & 0.54 &-&-\\
      {caps5trauma1related\_e03}          & 155                        & 50.3                     & 51.6 & 0.93                       & 0.86 &  -0.40                     & 0.17 &-&-\\
      {dsm5capscrite04trauma1\_distress}  & 252                        & 63.1                     & 52.8 & 0.85                       & 1.05 &   -0.03                    & 0.60 &-&-\\
      {caps5trauma1related\_e04}          & 117                        & 50.4                     & 53.0 & 0.99                       & 0.88 &  -0.55                     & 0.13 &-&-\\
      {dsm5capscrite05trauma1\_distress}  & 256                        & 59.8                     & 53.5 & 0.81                       & 0.99 &  -0.13                     & 0.58 &-&-\\
      {caps5trauma1related\_e05}          & 161                        & 57.8                     & 41.6 & 1.09                       & 0.99 &  -0.79                     & 0.51 &-&-\\
      {dsm5capscrite06trauma1\_distress}  & 256                        & 53.5                     & 52.7 & 1.02                       & 1.06 &  0.09                     & 0.37 &-&-\\
      {caps5trauma1related\_e06}          & 181                        & 63.0                     & 38.7 & 1.0                        & 10.2 &   -0.67                    & 0.37 &-&-\\
      {dsmiv\_future\_frequency\_current} & 251                        & 80.1                     & 48.6 & 0.81                       & 6.02 &  0.40                     & 0.80 &-&-\\
      {dsmiv\_future\_intens\_current}    & 246                        & 69.1                     & 40.2 & 0.93                       & 1.77 &  0.61                     & 0.90 &-&-\\
      {dsm5capscritg\_trauma1\_distress}  & 228                        & 53.9                     & 43.4 & 1.08                       & 1.39 &  0.35                     & 0.69 &-&-\\
      {dsm5capscritg\_trauma1\_impair}    & 226                        & 51.8                     & 42.0 & 0.93                       & 1.27 &  -0.28                     & 0.57 &-&-\\
      {dsm5capscritg\_trauma1\_fx}        & 205                        & 54.1                     & 29.3 & 1.10                       & 1.56 &   -0.04                    & 0.81 &-&-\\
      {dsm5depersonalization\_sev}        & 255                        & 67.5                     & 52.2 & 0.80                       & 1.25 &  -0.08                     & 0.49 &-&-\\
      {caps5trauma1related\_diss01}       & 76                         & 53.9                     & 31.6 & 1.16                       & 1.26 &  0.31                     & 0.19 &-&-\\
      {dsm5derealization\_sev}            & 249                        & 63.1                     & 30.9 & 0.98                       & 1.74 & 0.20                      & 0.88 &-&-\\
      {caps5trauma1related\_diss02}       & 70                         & 55.7                     & 27.1 & 1.11                       & 1.25 &   -0.03                    & 0.53 &-&-\\
      \hline
      \multicolumn{10}{c}{\bf \texttt{Category} Variable}\\
      \hline
      lbi\_a1                        & 200                       & 70.0                    & 41.0 &-&-&-&-&-&-\\
      lbi\_student                   & 201                       & 95.0                    & 89.1 &-&-&-&-&-&-\\
      lbi\_c1a                       & 192                       & 57.8                    & 71.9 &-&-&-&-&-&-\\
      lbi\_c2                        & 1                         & 100                     & 100  &-&-&-&-&-&-\\
      thh\_medicalcond               & 206                       & 92.7                    & 88.8 &-&-&-&-&-&-\\
      thh\_tx\_curr\_yesno           & 215                       & 94.9                    & 80.9 &-&-&-&-&-&-\\
      thh\_tx\_yesno                 & 233                       & 89.7                    & 87.6 &-&-&-&-&-&-\\
      feedback\_helpful              & 79                        & 94.9                    & 89.9 &-&-&-&-&-&-\\
      thh\_txneed\_yesno             & 96                        & 92.7                    & 88.5 &-&-&-&-&-&-\\
      thh\_psychmed\_curr\_yesno     & 194                       & 92.3                    & 88.7 &-&-&-&-&-&-\\
      thh\_psychmed\_yesno           & 198                       & 95.5                    & 93.4 &-&-&-&-&-&-\\
      thh\_suicide\_yesno            & 236                       & 90.7                    & 77.1 &-&-&-&-&-&-\\
      thh\_suicide\_pw\_yesno        & 70                        & 94.3                    & 78.6 &-&-&-&-&-&-\\
      trauma1lifeeventscl            & 146                       & 61.6                    & 12.3 &-&-&-&-&-&-\\
      trauma1\_exposure\_type\_\_\_1 & 146                       & 77.4                    & 67.1 &-&-&-&-&-&-\\
      trauma1\_exposure\_type\_\_\_2 & 146                       & 77.4                    & 43.2 &-&-&-&-&-&-\\
      trauma1\_exposure\_type\_\_\_3 & 146                       & 67.8                    & 28.1 &-&-&-&-&-&-\\
      trauma1\_exposure\_type\_\_\_4 & 146                       & 65.8                    & 22.6 &-&-&-&-&-&-\\
      caps\_e1\_lt                   & 145                       & 62.1                    & 45.5 &-&-&-&-&-&-\\
      caps\_e1\_ltself               & 73                        & 64.4                    & 64.4 &-&-&-&-&-&-\\
      caps\_e1\_ltother              & 74                        & 41.9                    & 44.6 &-&-&-&-&-&-\\
      caps\_e1\_si                   & 146                       & 43.8                    & 39.7 &-&-&-&-&-&-\\
      caps\_e1\_siself               & 61                        & 54.1                    & 65.6 &-&-&-&-&-&-\\
      caps\_e1\_siother              & 61                        & 60.7                    & 29.5 &-&-&-&-&-&-\\
      caps\_e1\_tpi                  & 146                       & 54.1                    & 52.7 &-&-&-&-&-&-\\
      caps\_e1\_tpiself              & 79                        & 84.8                    & 75.9 &-&-&-&-&-&-\\
      caps\_e1\_tpiother             & 77                        & 49.4                    & 26.0 &-&-&-&-&-&-\\
      trauma1\_nomemory              & 145                       & 75.2                    & 44.8 &-&-&-&-&-&-\\
      dsm5caps\_critf\_cur1\_yesno   & 202                       & 78.7                    & 41.6 &-&-&-&-&-&-\\
      dsm5caps\_critf\_cur1\_c       & 198                       & 83.3                    & 87.9 &-&-&-&-&-&-\\
      \hline
      \multicolumn{10}{c}{\bf \texttt{Measure} Variable}\\
      \hline
      lbi\_a2a                        & 99                        & 68.7                    & 66.7 &-&-& 41.9                     & 45.5 &-&-\\
      trauma1age                      & 143                       & 63.6                    & 55.9 &-&-& 21.2                     & 31.7 &-&-\\
      dsm5capscritb01trauma1\_num     & 162                       & 63.6                    & 58.0 &-&-& 37.3                     & 52.9 &-&-\\
      dsm5capscritb02trauma1\_num     & 98                        & 74.5                    & 63.3 &-&-& 28.0                     & 52.8 &-&-\\
      dsm5capscritb03trauma1\_num     & 84                        & 72.6                    & 59.5 &-&-& 47.8                     & 76.5 &-&-\\
      dsm5capscritb04trauma1\_num     & 177                       & 62.1                    & 58.8 &-&-& 17.9                     & 42.5 &-&-\\
      dsm5capscritb05trauma1\_num     & 137                       & 59.1                    & 57.7 &-&-& 28.6                     & 50.0 &-&-\\
      dsm5capscritc01trauma1\_num     & 170                       & 59.4                    & 53.5 &-&-& 31.9                     & 54.4 &-&-\\
      dsm5capscritc02trauma1\_num     & 140                       & 63.6                    & 54.3 &-&-& 27.5                     & 54.7 &-&-\\
      dsm5capscritd01trauma1\_num     & 87                        & 50.6                    & 48.3 &-&-& 32.6                     & 82.2 &-&-\\
      dsm5capscritd02trauma1\_num     & 168                       & 76.2                    & 69.0 &-&-& 47.5                     & 59.6 &-&-\\
      dsm5capscritd03trauma1\_num     & 120                       & 65.0                    & 57.5 &-&-& 23.8                     & 43.1 &-&-\\
      dsm5capscritd04trauma1\_num     & 166                       & 72.3                    & 68.1 &-&-& 39.1                     & 45.3 &-&-\\
      dsm5capscritd05trauma1\_num     & 138                       & 65.9                    & 59.4 &-&-& 42.6                     & 46.4 &-&-\\
      dsm5capscritd06trauma1\_num     & 155                       & 69.7                    & 63.2 &-&-& 27.7                     & 40.4 &-&-\\
      dsm5capscritd07trauma1\_num     & 140                       & 61.4                    & 60.0 &-&-& 40.7                     & 53.6 &-&-\\
      dsm5capscrite01trauma1\_num     & 135                       & 65.9                    & 62.2 &-&-& 21.7                     & 54.9 &-&-\\
      dsm5capscrite02trauma1\_num     & 61                        & 83.6                    & 68.9 &-&-& 80.0                     & 89.5 &-&-\\
      dsm5capscrite03trauma1\_num     & 159                       & 73.0                    & 67.3 &-&-& 37.2                     & 34.6 &-&-\\
      dsm5capscrite04trauma1\_num     & 131                       & 68.7                    & 60.3 &-&-& 24.4                     & 50.0 &-&-\\
      dsm5capscrite05trauma1\_num     & 168                       & 69.0                    & 66.1 &-&-& 21.2                     & 35.1 &-&-\\
      dsm5capscrite06trauma1\_num     & 184                       & 72.8                    & 61.4 &-&-& 40.0                     & 31.0 &-&-\\
      dsmcaps\_critf\_cur1\_nummonths & 191                       & 49.7                    & 22.5 &-&-& 60.4                     & 81.8 &-&-\\
      dsm5caps\_critf\_cur1\_b        & 181                       & 35.7                    & 22.0 &-&-& 17.9                     & 19.0 &-&-\\
      dsm5depersonalization\_num      & 84                        & 59.5                    & 51.2 &-&-& 32.4                     & 65.9 &-&-\\
      dsm5derealization\_num          & 3                         & 100                     & 33.3 &-&-& 0.00                     & 50.0 &-&-\\
      \hline
      \multicolumn{10}{c}{\bf \texttt{Notes} Variable}\\
      \hline
      life\_base\_typicalday       & 203                       &-&-&-&-&-&-& 42.0                       & 50.8 \\
      thh\_medicalcond\_desc       & 100                       &-&-&-&-&-&-& 56.8                       & 80.1 \\
      thh\_tx\_curr\_descr         & 59                        &-&-&-&-&-&-& 53.6                       & 73.8 \\
      thh\_tx\_descr               & 135                       &-&-&-&-&-&-& 44.0                       & 57.4 \\
      dx\_knowledge                & 33                        &-&-&-&-&-&-& 59.4                       & 48.5 \\
      dx\_lackknowledge            & 20                        &-&-&-&-&-&-& 60.7                       & 37.9 \\
      feedback\_info               & 66                        &-&-&-&-&-&-& 75.1                       & 48.1 \\
      thh\_txneed\_desc            & 45                        &-&-&-&-&-&-& 59.7                       & 49.4 \\
      thh\_psychmed\_descr         & 89                        &-&-&-&-&-&-& 40.4                       & 59.0 \\
      thh\_suicide\_desc           & 73                        &-&-&-&-&-&-& 56.9                       & 67.0 \\
      thh\_suicide\_pw\_desc       & 13                        &-&-&-&-&-&-& 62.2                       & 62.3 \\
      critaprobenotes              & 143                       &-&-&-&-&-&-& 50.8                       & 37.4 \\
      trauma1whathappened          & 143                       &-&-&-&-&-&-& 42.7                       & 51.4 \\
      trauma1describe              & 24                        &-&-&-&-&-&-& 51.4                       & 48.9 \\
      \hline
      \multicolumn{10}{c}{\bf \texttt{Rule} Variable}\\
      \hline
      lbi\_e2                       & 215                        & 43.3                     & 37.7 & 0.94                       & 0.98 & 0.44                      & 0.43 &-&-\\
      caps\_e1\_crita               & 146                        & 91.8                     & 71.9 & 0.38                       & 0.97 &  0.83                     & -0.51 &-&-\\
      dsm5capscritb01trauma1        & 253                        & 62.8                     & 63.6 & 0.81                       & 0.90 &  -0.51                     & 0.28 &-&-\\
      dsm5capscritb02trauma1        & 250                        & 88.0                     & 70.0 & 0.53                       & 0.80 & -0.47                      & 0.47 &-&-\\
      dsm5capscritb03trauma1        & 246                        & 86.2                     & 63.4 & 0.54                       & 0.96 & 0.00                      & 0.82 &-&-\\
      dsm5capscritb04trauma1        & 255                        & 67.5                     & 60.0 & 0.86                       & 0.95 & -0.57                      & 0.25 &-&-\\
      dsm5capscritb05trauma1        & 241                        & 74.7                     & 69.7 & 0.73                       & 0.75 &  -0.18                     & 0.26 &-&-\\
      dsm5capscritc01trauma1        & 250                        & 55.2                     & 54.8 & 0.94                       & 0.97 &  -0.64                     & 0.36 &-&-\\
      dsm5capscritc02trauma1        & 242                        & 71.9                     & 61.2 & 0.83                       & 0.93 &  -0.29                     & 0.51 &-&-\\
      dsm5capscritd01trauma1        & 239                        & 81.2                     & 66.5 & 0.68                       & 1.02 & 0.24                      & 0.60 &-&-\\
      dsm5capscritd02trauma1        & 222                        & 62.6                     & 46.4 & 0.79                       & 1.09 &  -0.16                     & 0.83 &-&-\\
      dsm5capscritd03trauma1        & 246                        & 72.0                     & 72.0 & 0.85                       & 0.74 & -0.48                      & 0.36 &-&-\\
      dsm5capscritd04trauma1        & 251                        & 63.7                     & 62.2 & 0.94                       & 1.02 &  -0.08                     & 0.35 &-&-\\
      dsm5capscritd05trauma1        & 252                        & 59.9                     & 53.6 & 0.98                       & 1.00 & -0.19                      & 0.42 &-&-\\
      dsm5capscritd06trauma1        & 254                        & 55.9                     & 50.8 & 1.03                       & 1.10 &  -0.07                     & 0.57 &-&-\\
      dsm5capscritd07trauma1        & 255                        & 63.1                     & 60.0 & 0.85                       & 0.95 & -0.17                      & 0.59 &-&-\\
      dsm5capscrite01trauma1        & 255                        & 72.5                     & 51.4 & 0.69                       & 0.90 & 0.03                      & 0.66 &-&-\\
      dsm5capscrite02trauma1        & 250                        & 90.4                     & 76.8 & 0.38                       & 0.73 &  0.75                     & 0.90 &-&-\\
      dsm5capscrite03trauma1        & 220                        & 57.3                     & 58.6 & 0.91                       & 0.96 &  0.09                     & 0.43 &-&-\\
      dsm5capscrite04trauma1        & 250                        & 75.6                     & 72.0 & 0.71                       & 0.79 &  -0.08                     & 0.63 &-&-\\
      dsm5capscrite05trauma1        & 254                        & 65.7                     & 67.7 & 0.80                       & 0.77 &  -0.36                     & 0.54 &-&-\\
      dsm5capscrite06trauma1        & 254                        & 55.9                     & 52.8 & 1.05                       & 0.96 & 0.05                      & 0.27 &-&-\\
      dsmcaps\_critf\_admin         & 28                         & 75.0                     & 100  & 0.50                       & 0.00 & -1.00                      & -1.00 &-&-\\
      dsm5depersonalization         & 246                        & 85.4                     & 64.2 & 0.61                       & 0.78 &  -0.06                     & 0.70 &-&-\\
      dsm5derealization             & 243                        & 75.3                     & 39.1 & 0.69                       & 1.14 &  0.27                     & 0.76 &-&-\\
      dsm5capsglobalvalidtrauma1    & 255                        & 63.5                     & 63.5 & 0.84                       & 0.84 & -1.00                      & -1.00 &-&-\\
      dsm5capsglobalsevtrauma1      & 254                        & 44.1                     & 42.9 & 0.91                       & 0.97 & 0.21                      & 0.45 &-&-\\
    \end{supertabular}
    \label{tab:var_results}
  \end{center}
}
\twocolumn

\end{document}